\documentclass[runningheads]{llncs}

\usepackage[T1]{fontenc}
\usepackage{graphicx,verbatim}
\usepackage{multirow}
\usepackage{booktabs}
\usepackage{graphicx}
\usepackage{xcolor}
\usepackage{amsmath}
\usepackage{amssymb}
\usepackage{caption}

\begin{document}

\title{DC-Seg: Disentangled Contrastive Learning for Brain Tumor Segmentation with Missing Modalities}
\author{Haitao Li \and Ziyu Li \and Yiheng Mao \and Zhengyao Ding \and Zhengxing Huang}
\authorrunning{H. Li et al.}
\institute{Zhejiang University \\
\email{lihaitao@zju.edu.cn, zhengxinghuang@zju.edu.cn}}
\maketitle          

\begin{abstract}
Accurate segmentation of brain images typically requires the integration of complementary information from multiple image modalities. However, clinical data for all modalities may not be available for every patient, creating a significant challenge. To address this, previous studies encode multiple modalities into a shared latent space. While somewhat effective, it remains suboptimal, as each modality contains distinct and valuable information. In this study, we propose DC-Seg (Disentangled Contrastive Learning for Segmentation), a new method that explicitly disentangles images into modality-invariant anatomical representation and modality-specific representation, by using anatomical contrastive learning and modality contrastive learning respectively. This solution improves the separation of anatomical and modality-specific features by considering the modality gaps, leading to more robust representations. Furthermore, we introduce a segmentation-based regularizer that enhances the model’s robustness to missing modalities. Extensive experiments on the BraTS 2020 and a private white matter hyperintensity(WMH) segmentation dataset demonstrate that DC-Seg outperforms state-of-the-art methods in handling incomplete multimodal brain tumor segmentation tasks with varying missing modalities, while also demonstrate strong generalizability in WMH segmentation. The code is available at https://github.com/CuCl-2/DC-Seg.

\end{abstract}

\section{Introduction}
Accurate brain image segmentation is crucial for assessing disease progression and developing effective treatments. Brain MRI, with modalities like T1, T2, T1ce, and FLAIR, provides varying sensitivity to lesion regions depending on imaging parameters and protocols \cite{ding2021rfnet}. Joint learning across these multimodal images improves segmentation accuracy compared to single-modality approaches. Common methods involve concatenating images from different modalities \cite{kamnitsas2017efficient,zhou2018one,chen20193d} or integrating features in high-dimensional spaces \cite{fidon2017scalable,tseng2017joint,zhang2020exploring}. However, missing modalities due to protocol variations or patient factors pose a challenge.

Significant efforts have been made to address the challenges posed by missing modalities in practical scenarios, with existing solutions falling into three main categories. The first approach synthesizes missing modalities to complete the test set \cite{van2015does,shen2020multi,meng2024multi}. This involves training a generative model to generate missing modalities, but it often requires additional training and struggles when only one modality is available during inference. The second approach trains a dedicated model for each specific missing-modal scenario. Methods like \cite{hu2020knowledge,chen2021learning,wang2021acn,azad2022smu} distill knowledge from a multimodal teacher network to monomodal students at the image and pixel levels. Considering the varying sensitivities of lesion regions across modalities, GSS \cite{qiu2023scratch} selects a group leader for distillation. While these methods perform well when multiple modalities are missing, they incur high computational and memory costs, requiring $2^N-1$ models for $N$ modalities. The third approach attempts to handle all missing-modal situations with a single unified model, embedding all modalities into a shared latent space, followed by feature fusion for segmentation \cite{havaei2016hemis,chen2019robust}. RFNet \cite{ding2021rfnet} uses a region-aware fusion module to adaptively combine features from available modalities on different regions, while mmFormer \cite{zhang2022mmformer} leverages Transformer for long-range dependencies, and M\textsuperscript{3}AE \cite{liu2023m3ae} employs multimodal autoencoders to reduce model complexity by creating a unified latent representation.

While valuable, these studies often overlook modality gaps, failing to learn invariant feature representations across modalities, which impairs performance in missing-modality scenarios. To address this, some approaches \cite{chen2019robust,azad2022smu,yang2022d} decompose images into modality-invariant and modality-specific components, using invariant representations for segmentation. For instance, SMU-Net \cite{azad2022smu} posits that deeper network layers capture content representations, while shallower layers preserve style representations. Similarly, RobustSeg \cite{chen2019robust} decouples content and appearance codes by reconstructing images. D2Net \cite{yang2022d} learns modality-specific codes through contrastive learning applied to different MRI slices. 

In this study, we introduce bidirectional contrastive learning, complementing the traditional reconstruction task to achieve effective decoupling. Unlike previous models, our approach applies both anatomical and modality contrastive learning at the 3D MRI image level. This allows us to learn not only modality representations but also modality-invariant anatomical representations which are crucial for accurate segmentation. By performing contrastive learning on the full 3D image, we achieve more comprehensive feature extraction. Specifically, anatomical contrastive learning pulls features from the same individual across modalities closer, while pushing features from different individuals apart. Similarly, modality contrastive learning pulls features from the same modality across different individuals closer while pushing features from different modalities apart. Additionally, a segmentation-based regularizer is incorporated to further enhance the model’s robustness to incomplete modalities. 

We validate our method on the BRATS \cite{menze2014multimodal} dataset for multimodal brain tumor segmentation, achieving competitive performance in full-modality scenarios and superior robustness in missing-modality settings. Additionally, we demonstrate its generalizability on a private WMH segmentation dataset.

\begin{figure}[t]
    \centering
    \includegraphics[width=1.0\linewidth]{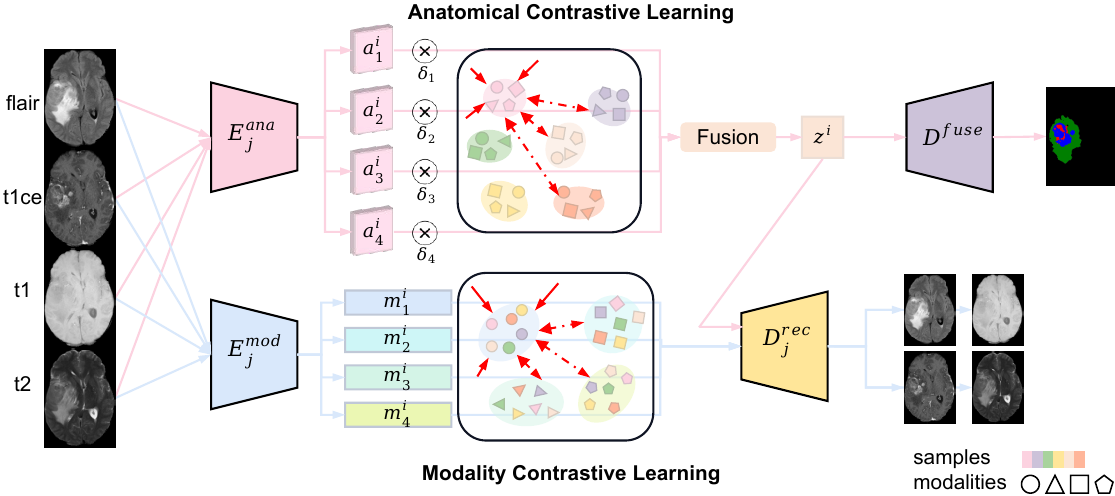}
    \caption{Overview of DC-Seg, which disentangles images from different modalities into anatomical and modality representations using bidirectional contrastive learning, and fuses modality-invariant anatomical representations for the downstream tumor segmentation task. For clarity in the figure, $D^{\text{sep}}$ is omitted.}
    \label{fig:method}
\end{figure}

\section{Method}
An overview of our proposed DC-Seg is shown in Fig. \ref{fig:method}. First, we decouple the multimodal inputs into modality-specific and modality-invariant anatomical representations using both the traditional reconstruction task and our proposed novel bidirectional contrastive learning approach. Next, we fuse the anatomical representations from different modalities for tumor segmentation. Additionally, a segmentation-based regularizer is introduced to prevent the model from becoming highly dependent on discriminative modalities (e.g., T1ce, FLAIR) for brain tumor recognition, which could lead to significant degradation in performance when these discriminative modalities are missing. The detailed learning process and network architecture are described below.

\subsection{Bidirectional Contrastive Learning}
In this part, we describe how images are disentangled into anatomical and modality-specific representations. Traditional methods \cite{lee2018diverse,ouyang2021representation,chen2019robust} achieve this by reconstructing images from fused anatomical representation and modality-specific representations across all modalities. However, these methods only ensure that the fused anatomical representation is well-learned and do not guarantee that the anatomical representation for each modality is modality-invariant, which can lead to performance degradation when modalities are missing. To address this issue, we introduce bidirectional contrastive learning to ensure that anatomical representations across all modalities are effectively learned and aligned.

Here’s how it works, given a batch of multimodal images \(\{x_j^i\}\), where \(i \in \{1, \dots, N\}\) indexes the samples and \(j \in \{1, \dots, M\}\) indexes the modalities, with \(M = 4\) in our brain tumor segmentation task, each modality \(x_j^i\) from sample \(x^i\) is passed through its respective anatomical encoder \(E_j^{\text{ana}}\) and modality encoder \(E_j^{\text{mod}}\) to obtain corresponding disentangled anatomical representation \(a_j^i = E_j^{\text{ana}}(x_j^i)\) and modality representation \(m_j^i = E_j^{\text{mod}}(x_j^i)\). 
For the modality representation, we follow the common practice in \cite{lee2018diverse}, representing it as an 8-bit vector \( m_j^i \in \mathbb{R}^C\), while the anatomical representation is encoded as multichannel 3D feature maps \( a_j^i \in \mathbb{R}^{C \times d \times d \times d} \). Intuitively, we aim to align different modalities of the same sample to share a common anatomical representation, which is essential for downstream tasks like brain tumor segmentation. For modality-specific representations, we ensure that images from the same modality are closely aligned. To achieve these objectives, we introduce bidirectional contrastive learning.

In anatomical contrastive learning, given an anchor image \( x_j^i \), the positive samples are images from the same subject, denoted as \( x_{j'}^{i'} \), \( i' = i \) while the negative samples are from different subjects (\( i' \neq i \)). Since each anchor image has multiple positive samples, we do not use the softmax-based contrastive loss as in CLIP \cite{radford2021learning}. Instead, we employ a sigmoid-based loss similar to \cite{zhai2023sigmoid}. Formally, the anatomical contrastive loss \( L_{\text{ana}} \) is defined as:

\begin{equation}
\mathcal{L}_{\text{ana}} = -\frac{1}{(N \cdot M)^2} \sum_{i,j,i',j'} \log \frac{1}{1 + e^{f(i, i') \cdot \left( -t \cdot \text{SSIM}(a_j^i, a_{j'}^{i'}) \right)}}
\label{eq:ana_loss}
\end{equation}

\begin{equation}
f(i, i') =
\begin{cases}
1, & \text{if } i = i', \\
-1, & \text{if } i \neq i'
\end{cases}
\end{equation}

where a batch containing \(N \times M\) images results in \((N \times M)^2\) pairs. \( f(i, i') \) is used to determine whether two images belong to the same sample. The temperature scaling factor \( t \) controls the sharpness of the distribution, influencing the model’s sensitivity to positive and negative pairs. The \( \text{SSIM}(a_{j}^i, a_{j'}^{i'}) \) defined below represents the channel-wise mean of structural similarity between the feature maps \( a_{j,c}^i \) and \( a_{j',c}^{i'} \) for each channel. The constants \( C_1 \) and \( C_2 \) are small values introduced to prevent division by zero and stabilize the computation.

\begin{equation}
\text{SSIM}(a_{j}^i, a_{j'}^{i'}) = \frac{1}{C} \sum_{c=1}^{C} \frac{(2\mu_{a_{j,c}^i} \mu_{a_{j',c}^{i'}} + C_1)(2\sigma_{a_{j,c}^i, a_{j',c}^{i'},} + C_2)}{(\mu_{a_{j,c}^i}^2 + \mu_{a_{j',c}^{i'}}^2 + C_1)(\sigma_{a_{j,c}^i}^2 + \sigma_{a_{j',c}^{i'}}^2 + C_2)}
\end{equation}

Similar to the anatomical contrastive loss in Eq. \ref{eq:ana_loss}, the modality contrastive loss \( \mathcal{L}_{\text{mod}} \) is defined as Eq. \ref{eq:mod_loss}, with a key distinction: images from the same modality are treated as positive pairs (i.e., \( j = j' \)), while images from different modalities are treated as negative pairs. Since the modality representation \( m_{i,j} \) is an 8-bit vector, cosine similarity is used instead of SSIM. The similarity is defined as: $ \text{sim}(m_j^i, m_{j'}^{i'}) = \frac{m_j^i \cdot m_{j'}^{i'}}{\|m_j^i\|_2 \|m_{j'}^{i'}\|_2} $.

\begin{equation}
\mathcal{L}_{\text{mod}} = -\frac{1}{(N \cdot M)^2} \sum_{i,j,i',j'} \log \frac{1}{1 + e^{f(j, j') \cdot \left( -t \cdot \text{sim}(m_j^i, m_{j'}^{i'}) \right)}}
\label{eq:mod_loss}
\end{equation}

In addition to the bidirectional contrastive learning discussed above, we adhere to the assumption that, for successful disentanglement, the obtained anatomical representation should be re-renderable into the original image when paired with the modality representation of any given modality \cite{lee2018diverse}. Specifically, we fuse the anatomical representations from different modalities to obtain \( z^i \) following \cite{ding2021rfnet}, and then reconstruct the image using a set of modality-specific decoders, \(\{D_j^{\text{rec}}\}\), given \( z^i \) and the modality representation \( m_j^i \). The loss function is defined below, where we use the L1-norm to prevent image blurring. A Bernoulli indicator \( \delta_i \) is employed to enhance the robustness of the content representation \( z \) to missing data, with modality dropout applied in the latent space by randomly setting \( \delta_i \) to 0.

\begin{equation}
\mathcal{L}_{\text{rec}} = \sum_{i=1}^{N} \sum_{j=1}^{M} \| D_j^{\text{rec}}(z^i, m_j^i) - x_j^i \|_1, \quad \text{where} \quad z^i = \mathcal{F}(\delta_1 a_1^i,\delta_2 a_2^i, \ldots, \delta_M a_M^i),
\end{equation}

The final disentanglement loss is defined as follows.

\begin{equation}
\mathcal{L}_{\text{disentangle}} = \mathcal{L}_{\text{ana}} + \mathcal{L}_{\text{mod}} + \mathcal{L}_{\text{rec}}
\end{equation}

\subsection{Learning Process}
Due to the high sensitivity of certain discriminative modalities (e.g., T1ce, FLAIR) to specific tumor regions, the model tends to depend on these modalities for segmentation, resulting in significant performance degradation when they are unavailable. Therefore, it is critical to encourage the model to segment based on all modalities. To achieve this, we introduce a segmentation-based regularizer like \cite{ding2021rfnet,zhang2022mmformer}. Specifically, we use a weight-shared decoder $D^{\text{sep}}$ to segment based on every single modality separately. The corresponding weighted cross-entropy loss and Dice loss are used as regularization terms, expressed as:
\begin{equation}
\mathcal{L}_{\text{reg}} = \sum_{i=1}^{N} \sum_{j=1}^{M} \left( \mathcal{L}_{\text{WCE}}(D^{\text{sep}}(a_j^i), y^i) + \mathcal{L}_{\text{DL}}(D^{\text{sep}}(a_j^i), y^i) \right),
\end{equation}

As illustrated in Fig. \ref{fig:method}, the fused anatomical feature \( z^i \) is used to predict the final segmentation mask through \( D^{\text{fuse}} \). The weighted cross-entropy loss and Dice loss are employed to align the predictions with the corresponding ground-truth segmentation maps, as expressed below:
\begin{equation}
\mathcal{L}_{\text{seg}} = \sum_{i=1}^{N} \left( \mathcal{L}_{\text{WCE}}(D^{\text{fuse}}(z^i), y^i) + \mathcal{L}_{\text{DL}}(D^{\text{fuse}}(z^i), y^i) \right),
\end{equation}

Therefore, the overall loss of our DC-Seg is defined below, with \( \alpha \) as a hyperparameter for the tradeoff.
\begin{equation}
\mathcal{L} = \mathcal{L}_{\text{seg}} + \mathcal{L}_{\text{reg}} + \alpha \mathcal{L}_{\text{disentangle}}
\end{equation}

\section{Experiments and Results}
\textbf{Datasets and Implementation.}
The experiments are conducted using the BraTS 2020 dataset \cite{menze2014multimodal} and a private white matter hyperintensity segmentation dataset(SAHZU-WMH) from The Second Affiliated Hospital, Zhejiang University School of Medicine.

The BraTS 2020 dataset includes 369 multi-contrast MRI scans across four modalities (T1, T1c, T2, FLAIR), with tumor subregions: whole tumor, tumor core, and enhancing tumor. All volumes are skull-stripped, co-registered to a common anatomical template, resampled to 1mm³ isotropic resolution, and normalized to zero mean and unit variance within the brain tissue. Patches of size 112 × 112 × 112 are randomly cropped and used as input to the network during training.

The SAHZU-WMH dataset includes 41 patients with cognitive impairment, each with two follow-up MRI scans (average interval of 473 days), totaling 80 multi-modal scans (FLAIR and T1) after excluding one missing follow-up and one incorrect annotation. White matter hyperintensity regions are annotated. Preprocessing follows the same steps as BraTS 2020, with patches of size 128×128×128. The dataset is split into 60 scans from 31 individuals for training and 20 scans from 10 individuals for testing, ensuring no overlap between subjects in the training and test sets.

Random flips, cropping, and intensity shifts are applied for data augmentation. The network is trained using the Adam optimizer with an initial learning rate of 0.0002 for 500 epochs with batch size 2. The hyperparameters are set as: \( \alpha = 0.4 \).

\begin{table*}[t!]
    \centering
    \caption{Results of the proposed method and state-of-the-art unified models on BraTS 2020 dataset. Dice similarity coefficient is employed for evaluation with every combination of modality settings. $\bullet$ and $\circ$ denote available and missing modalities, respectively. }
    \label{tab:incomplete}
    \resizebox{\textwidth}{!}{
    \begin{tabular}{cccc|ccccc|ccccc|ccccc}
    \toprule
    \multicolumn{4}{c|}{\textbf{Modalities}} & \multicolumn{5}{c|}{\textbf{Complete}} & \multicolumn{5}{c|}{\textbf{Core}} & \multicolumn{5}{c}{\textbf{Enhancing}} \\ 
    \midrule
    F & T1 & T1c & T2 & RobustSeg & RFNet & mmFormer & M\textsuperscript{3}AE & \textbf{DC-Seg} & RobustSeg & RFNet & mmFormer & M\textsuperscript{3}AE & \textbf{DC-Seg} & RobustSeg & RFNet & mmFormer & M\textsuperscript{3}AE & \textbf{DC-Seg} \\
    \midrule
    $\circ$ & $\circ$ & $\circ$ & $\bullet$ & 82.20 & 86.05 & 85.51 & 86.10 & \textbf{86.72} & 61.88 & 71.02 & 63.36 & \textbf{71.80} & 70.88 & 36.46 & 46.29 & \textbf{49.09} & 47.10 & 47.76 \\
    $\circ$ & $\circ$ & $\bullet$ & $\circ$ & 71.39 & 76.77 & 78.04 & 78.90 & \textbf{79.54} & 76.68 & 81.51 & 81.51 & 83.60 & \textbf{84.62} & 67.91 & 74.85 & 78.30 & 73.60 & \textbf{78.90} \\
    $\circ$ & $\bullet$ & $\circ$ & $\circ$ & 71.41 & 77.16 & 76.24 & \textbf{79.00} & 78.47 & 54.30 & 66.02 & 63.23 & \textbf{69.40} & 66.63 & 28.99 & 37.30 & 37.62 & 40.40 & \textbf{42.19} \\
    $\bullet$ & $\circ$ & $\circ$ & $\circ$ & 82.87 & 87.32 & 86.54 & \textbf{88.00} & 87.80 & 60.72 & 69.19 & 64.60 & 68.70 & \textbf{71.27} & 34.68 & 38.15 & 36.68 & 40.20 & \textbf{41.66} \\
    $\circ$ & $\circ$ & $\bullet$ & $\bullet$ & 85.97 & 87.74 & 87.52 & 87.10 & \textbf{88.17} & 82.44 & 83.45 & 82.69 & 85.60 & \textbf{86.34} & 71.42 & 75.93 & 77.20 & 76.00 & \textbf{80.43} \\
    $\circ$ & $\bullet$ & $\bullet$ & $\circ$ & 76.84 & 81.12 & 80.70 & 80.10 & \textbf{82.22} & 80.28 & 83.40 & 82.81 & 83.80 & \textbf{85.18} & 70.11 & 78.01 & \textbf{81.71} & 75.30 & 79.25 \\
    $\bullet$ & $\bullet$ & $\circ$ & $\circ$ & 88.10 & 89.73 & 88.76 & 89.60 & \textbf{90.01} & 68.18 & 73.07 & 71.76 & 72.80 & \textbf{74.50} & 39.67 & 40.98 & 42.98 & 43.70 & \textbf{46.90} \\
    $\circ$ & $\bullet$ & $\circ$ & $\bullet$ & 85.53 & 87.73 & 86.94 & 87.30 & \textbf{88.09} & 66.46 & \textbf{73.13} & 67.76 & 72.90 & 73.09 & 39.92 & 45.65 & 49.12 & 48.70 & \textbf{50.19} \\
    $\bullet$ & $\circ$ & $\circ$ & $\bullet$ & 88.09 & 89.87 & 89.49 & 90.10 & \textbf{90.32} & 68.20 & 74.14 & 70.34 & 74.30 & \textbf{75.11} & 42.19 & 49.32 & 49.06 & 47.10 & \textbf{51.32} \\
    $\bullet$ & $\circ$ & $\bullet$ & $\circ$ & 87.33 & 89.89 & 89.31 & 89.50 & \textbf{89.99} & 81.85 & 84.65 & 83.79 & 85.50 & \textbf{85.90} & 70.78 & 76.67 & 79.44 & 75.90 & \textbf{80.28} \\
    $\bullet$ & $\bullet$ & $\bullet$ & $\circ$ & 88.87 & \textbf{90.69} & 89.79 & 89.60 & 90.65 & 82.76 & 85.07 & 84.44 & 85.60 & \textbf{86.29} & 71.77 & 76.81 & 80.65 & 76.30 & \textbf{81.41} \\
    $\bullet$ & $\bullet$ & $\circ$ & $\bullet$ & 89.24 & 90.60 & 89.83 & 90.20 & \textbf{90.77} & 70.46 & 75.19 & 72.42 & 74.40 & \textbf{75.53} & 43.90 & 49.92 & 50.08 & 48.20 & \textbf{52.05} \\
    $\bullet$ & $\circ$ & $\bullet$ & $\bullet$ & 88.68 & \textbf{90.68} & 90.49 & 90.50 & 90.62 & 81.89 & 84.97 & 83.94 & 85.80 & \textbf{86.21} & 71.17 & 77.12 & 78.73 & 77.40 & \textbf{79.42} \\
    $\circ$ & $\bullet$ & $\bullet$ & $\bullet$ & 86.63 & 88.25 & 87.64 & 87.40 & \textbf{88.73} & 82.85 & 83.47 & 83.66 & 85.80 & \textbf{86.49} & 71.87 & 76.99 & 77.34 & 78.00 & \textbf{81.66} \\
    $\bullet$ & $\bullet$ & $\bullet$ & $\bullet$ & 89.47 & \textbf{91.11} & 90.54 & 90.40 & 90.95 & 82.87 & 85.21 & 84.61 & 86.20 & \textbf{86.46} & 71.52 & 78.00 & 79.92 & 77.50 & \textbf{81.52} \\
    \midrule
    \multicolumn{4}{c}{Average} & 84.17 & 86.98 & 86.49 & 86.90 & \textbf{87.54} & 73.45 & 78.23 & 76.06 & 79.10 & \textbf{79.63} & 55.49 & 61.47 & 63.19 & 61.70 & \textbf{65.00} \\
    \bottomrule
    \end{tabular}
    }
\end{table*}

\begin{figure}[h]
    \centering
    \includegraphics[width=1.0\linewidth]{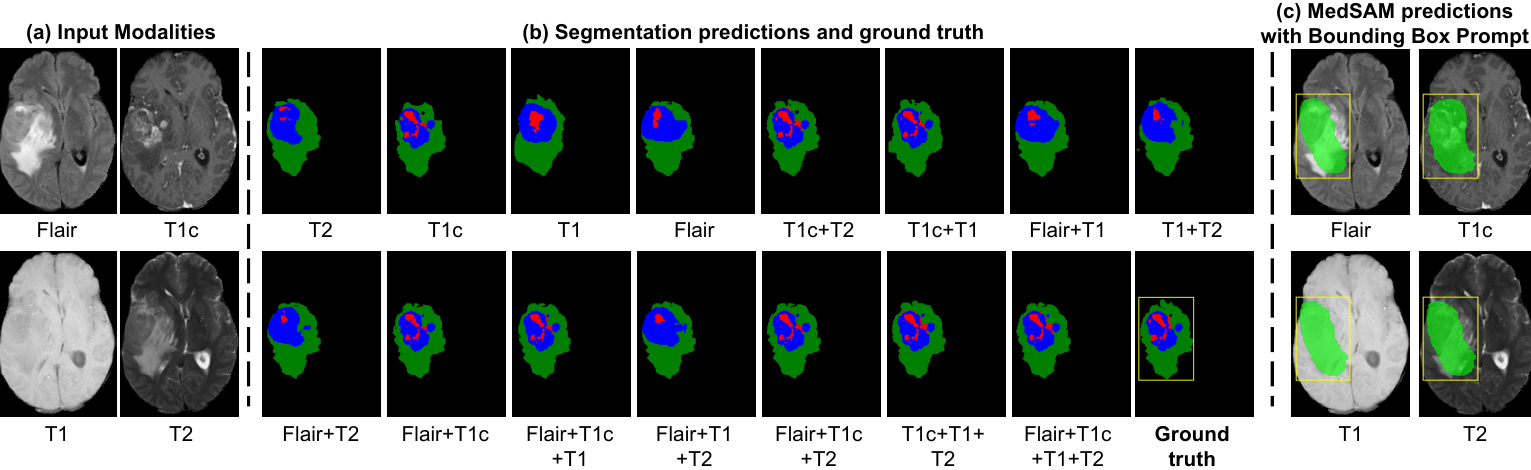}
    \caption{Visualization of the input modalities, our predicted segmentation maps, and MedSAM prediction with bounding box prompt.}
    \label{fig:res_visual}
\end{figure}

\textbf{Performance of Incomplete Multimodal Segmentation.}
We evaluate the robustness of our method for incomplete multimodal segmentation. The absence of a modality is simulated by setting \( \delta_i \) to zero. We compare our method against state-of-the-art approaches, namely RobustSeg \cite{chen2019robust}, RFNet \cite{ding2021rfnet}, mmFormer \cite{zhang2022mmformer}, M\textsuperscript{3}AE \cite{liu2023m3ae} and GSS\cite{qiu2023scratch}. For a fair comparison, we use the same data split as in \cite{ding2021rfnet} and directly reference the results. As shown in Table \ref{tab:incomplete}, our method significantly outperforms the state-of-the-art methods in the segmentation of all three tumor parts across most of the 15 possible modality combinations. Moreover, our approach surpasses the large pre-trained M\textsuperscript{3}AE model \cite{liu2023m3ae} (Complete: 86.9, Core: 79.1, Enhancing: 61.7) and achieves comparable performance to the dedicated method GSS \cite{qiu2023scratch} (Complete: 87.33, Core: 79.38, Enhancing: 65.54), which requires 15 models for all modalities combination. Figure \ref{fig:res_visual} illustrates that our method effectively segments brain tumors across various missing modality scenarios. Additionally, we also tested the medical foundation model MedSAM \cite{ma2024segment} for segmenting the complete tumor. However, even when provided with a bounding box as a prompt, MedSAM fails to clearly delineate the tumor's contour, further demonstrating the continued significance of our approach, even in the era of prevalent foundation models.

\begin{figure}
    \centering
    \includegraphics[width=0.95\linewidth]{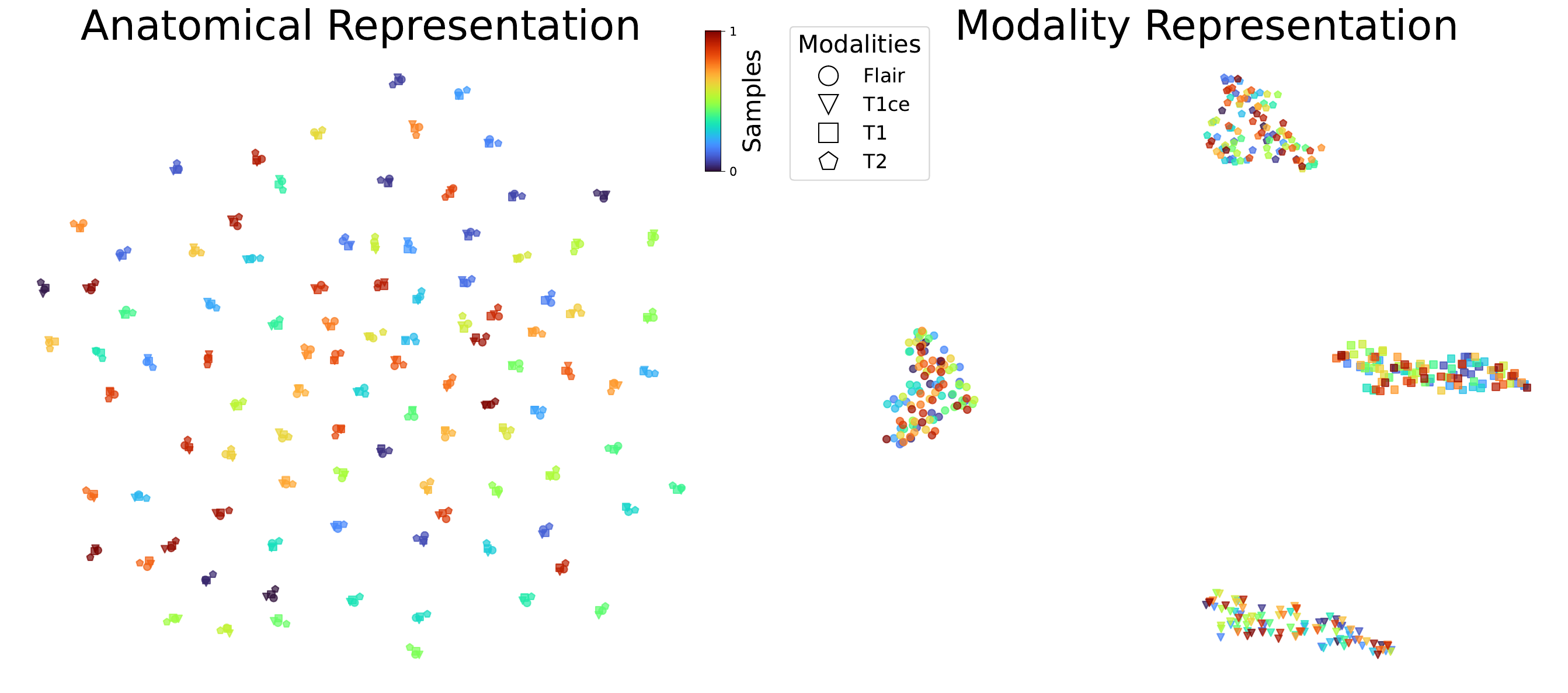}
    \caption{Visualization of anatomical and modality representations on the unseen test set of BRATS.}
    \label{fig:tsne_visual}
\end{figure}

\textbf{Disentanglement Visualization}
Figure \ref{fig:tsne_visual} visualizes the anatomical and modality representations on the unseen BRATS test set. After anatomical contrastive learning, modality-invariant anatomical representations are effectively aligned for each modality, improving the model's robustness to missing modalities. Additionally, modality-specific representations are successfully learned.

\textbf{Ablation Study.}
We investigate the effectiveness of anatomical contrastive learning, modality contrastive learning, the reconstruction task, and the regularizer as key components of our method. To assess the contribution of each component, we evaluate the performance of DC-Seg with each component excluded. In Table \ref{tab:ablation}, we compare the performance of these variants against the full DC-Seg model, measured by the Dice Similarity Coefficient (DSC), averaged over the 15 possible combinations of input modalities. The results show that each component contributes to performance improvement across all tumor subregions.

\begin{table}[h!]
\centering
\begin{minipage}{0.48\textwidth}
    \centering
    \caption{Ablation study. \scriptsize \textit{Ana and Mod: Anatomical and modality contrastive learning, Rec: reconstruction, Reg: Regularizer}}
    \resizebox{\textwidth}{!}{
        \begin{tabular}{cccc|ccc}
        \toprule
        Ana & Mod & Rec & Reg & Complete & Core & Enhancing\\ 
        \midrule
        \checkmark & \checkmark & \checkmark & \checkmark & 87.54 & 79.63 & 65.00\\
        $\times$ & $\times$ & \checkmark & \checkmark & 85.64 & 76.26 & 61.83 \\
        $\times$ & \checkmark & \checkmark & \checkmark & 85.81 & 77.15 & 62.37 \\
        \checkmark & $\times$ & \checkmark & \checkmark & 86.31 & 77.53 & 63.41 \\
        \checkmark & \checkmark & $\times$ & \checkmark & 86.58 & 77.54 & 64.48\\
        \checkmark & \checkmark & \checkmark & $\times$ & 86.38 & 76.58 & 63.40\\
        \bottomrule
        \end{tabular}
    }
    \label{tab:ablation}
\end{minipage}
\hfill
\begin{minipage}{0.48\textwidth}
    \centering
    \caption{Results of WMH Segmentation. Dice similarity coefficient is employed for evaluation}
    \resizebox{\textwidth}{!}{
        \begin{tabular}{cc|cccc}
        \toprule
        \multicolumn{2}{c|}{\textbf{Modalities}} & \multicolumn{4}{c}{\textbf{Dice scores}} \\ 
        \midrule
        \textbf{Flair} & \textbf{T1} & \textbf{RobustSeg} & \textbf{RFNet} & \textbf{mmFormer} & \textbf{Ours} \\ 
        \midrule
        $\circ$ & $\bullet$ & 55.56 & 58.26 & 53.91 & \textbf{60.42} \\ 
        $\bullet$ & $\circ$ & 81.69 & 82.35 & 77.29 &  \textbf{83.03} \\ 
        $\bullet$ & $\bullet$ & 82.24 & 82.66 & 77.98 & \textbf{82.84} \\ 
        \multicolumn{2}{c|}{\textbf{Average}} & 73.16 & 74.42 & 69.73 & \textbf{75.43}\\
        \bottomrule
        \end{tabular}
    }
    \label{tab:wmh}
\end{minipage}
\end{table}

\textbf{Performance of WMH Segmentation.}  
We evaluate the performance of our method for WMH segmentation across various modality combinations. The results, shown in Table \ref{tab:wmh}, highlight the effectiveness of our approach in comparison to state-of-the-art methods. Our method consistently outperforms all other methods across all modality combinations. Notably, in the case where only the T1 modality is available—where white matter hyperintensities are particularly challenging to discern—our approach still achieves the highest performance. These results emphasize the superior performance of our approach in WMH segmentation besides tumor segmentation, demonstrating its generalizability.

\section{Conclusion}
We propose DC-Seg, a novel multimodal segmentation framework that jointly uses anatomical contrastive learning and modality contrastive learning to decompose images into modality-invariant anatomical representations and modality-specific representations. We demonstrate the superiority of DC-Seg through extensive experiments on both the BraTS 2020 brain tumor dataset and a private white matter hyperintensity segmentation dataset, achieving state-of-the-art results in full-modality scenarios and outperforming existing methods in missing-modality conditions.

\bibliographystyle{splncs04}
\bibliography{mybibliography}

\end{document}